\documentclass[conference]{IEEEtran}

\IEEEoverridecommandlockouts
\usepackage[utf8]{inputenc} 
\usepackage{multirow}
\usepackage{longtable}
\usepackage{cite}
\usepackage{amsmath,amssymb,amsfonts}
\usepackage{algorithmic}
\usepackage{graphicx}
\usepackage{xcolor}
\usepackage{textcomp}
\def\BibTeX{{\rm B\kern-.05em{\sc i\kern-.025em b}\kern-.08em
    T\kern-.1667em\lower.7ex\hbox{E}\kern-.125emX}}
\begin{document}

\title{Autonomous Navigation and Collision Avoidance for Mobile Robots: Classification and Review\\
}

\author{
\IEEEauthorblockN{Marcus V. L. Carvalho}
\IEEEauthorblockA{\textit{PPGEE} \\
\textit{Poli USP }\\
São Paulo, Brazil \\
marcusvini178@usp.br}
\and
\IEEEauthorblockN{Roberto Simoni}
\IEEEauthorblockA{\textit{POSECM - UFSC} \\
\textit{UFSC}\\
Joinville, Brazil \\
roberto.simoni@ufsc.br}
\and
\IEEEauthorblockN{Leopoldo R. Yoshioka}
\IEEEauthorblockA{\textit{PPGEE} \\
\textit{ Poli USP }\\
São Paulo, Brazil \\
leopoldo.yoshioka@usp.br}
}

\maketitle

\begin{abstract}
This paper introduces a novel classification for  Autonomous Mobile Robots (AMRs), into three phases and five steps, focusing on autonomous collision-free navigation. Additionally, it presents the main methods and widely accepted technologies for each phase of the proposed classification. The purpose of this classification is to facilitate understanding and establish connections between the independent input variables of the system (hardware, software) and autonomous navigation. By analyzing well-established technologies in terms of sensors and methods used for autonomous navigation, this paper aims to provide a foundation of knowledge that can be applied in future projects of mobile robots.
\end{abstract}

\begin{IEEEkeywords}
Autonomous Mobile Robots, Navigation, Sensors, Methods, Obstacle Avoidance
\end{IEEEkeywords}

\section{Introduction}

Autonomous Mobile Robots (AMRs) are becoming increasingly essential in various sectors. They assist humans in performing complex, hazardous, or repetitive tasks. Initially created to improve productivity and safety in industrial settings, their scope has significantly broadened. From initially focusing on path planning for industrial manipulators \cite{khatib1986real}, AMRs now use advanced algorithms to navigate without collisions. This expansion has allowed them to operate in diverse and dynamic environments beyond just industrial settings \cite{panigrahi2022localization,sanchez2021path}. 

Despite considerable advancements, existing navigational strategies for Autonomous Mobile Robots (AMRs) often remain focused on specific domains: terrestrial, aerial, and aquatic. These strategies typically adopt layered approaches from perception to control, each tailored to distinct operational environments such as industrial settings \cite{macenski2020marathon2}, uneven terrains \cite{atas2021evaluation, atas2023navigating}, and underwater exploration \cite{lee2004intelligent, antonelli2001real}. All these applications suggest a lack of a unified framework that can seamlessly be integrated across all domains, a gap this paper aims to address. By adopting modular packages, the proposed classification enhances the reusability and interoperability of components, facilitating easier integration across all domains of autonomous navigation \cite{siegwart2011introduction, hajjaj2017bringing}.


This paper introduces a new, comprehensive classification system aimed at streamlining the various aspects of autonomous navigation. The system acts as a fundamental framework, organizing the intricate relationships between phases, modules, and layers. It improves the comprehension and execution of autonomous navigation strategies, offering clear insights, and ultimately offering a complete set of tools for practitioners to choose the best solution for a wide range of operational scenarios.


The paper is organized as follows: Section II presents the methodology and the process undertaken to develop our classification and review of components and technologies. Section III outlines the unified classification. Section IV discusses technological integrations and their applicability within various domains. Section V explores potential future directions and innovations, and Section VI concludes with key findings and implications for future research.

\section{Methodology for Literature Review}
\label{subsec:Methodology}

This study applied a systematic literature review approach to summarize the existing classifications and technologies in the field of autonomous navigation. In parallel with the analytical rigor demonstrated by \cite{chen2022milestones} in the autonomous vehicle domain. We aim to systematically identify and categorize significant contributions across the spectrum of autonomous mobile robots.

The objectives were to identify key classifications of autonomous navigation and analyze the integration of hardware, software, and the robot-environment dynamic.

The research criteria focused on:

\begin{itemize}
    \item Peer-reviewed papers that provided foundational insights and demonstrated long-term impact in the field.

    \item Clear frameworks for autonomous navigation, aiming to bridge foundational theories with contemporary advancements.
    
    \item Practical applications of technologies with significant developments in autonomous navigation.
\end{itemize}


Searches were conducted across major databases such as IEEE Xplore, ScienceDirect, Web of Knowledge, and SpringerLink using keywords like "autonomous navigation," "robot classification," and "obstacle avoidance." This strategic approach facilitated the inclusion of seminal, consolidated, and cutting-edge contributions.

This literature review allows to propose a novel classification intended to bridge existing gaps and facilitate future research in the field of autonomous navigation.

\section{A Unified Classification for Diverse Autonomous Navigation Applications}

\label{sec:Phases}

\begin{figure}[htb]
\begin{center}
\includegraphics[width=\linewidth]{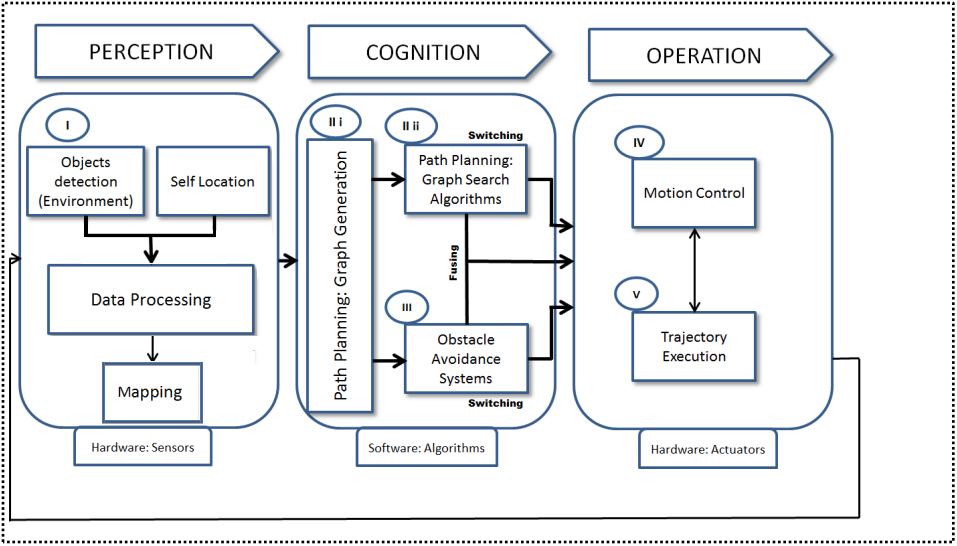}
\caption{Layers and phases of autonomous navigation.}
\label{fig:phasesAN} 
\end{center}
\end{figure}

This study has synthesized techniques, methodologies, and technologies for autonomous navigation, focusing on the dynamic interaction between hardware, software, and the robot environment. Based on this synthesis, we propose a unique classification system comprising three layers and five interconnected phases for autonomous navigation. The details of this classification system can be found on Figure \ref{fig:phasesAN}.

This classification interconnects the layers and phases as follows:

\begin{enumerate}
    \item Layer 1 - Perception
        \begin{enumerate}
            \item Phase I:  Environment Perception, Self Location, Data Processing, and Mapping.
        \end{enumerate}
    \item Layer 2 - Cognition
        \begin{enumerate}
            \item Phase II: Path Planning, including Graph Construction and Graph Search.
            \item Phase III Obstacle Avoidance and Trap landscape.
        \end{enumerate}
    \item Layer 3 - Operation
        \begin{enumerate}
            \item Phase IV Motion control,
            \item Phase V Path Execution
        \end{enumerate}
\end{enumerate}

The phases are not isolated but interconnected, covering the perception, cognition, and operation layers of mobile robotics. For instance, in the perception layer, mapping is integrated with data processing to generate a high-level environment representation, which is then used in the cognition layer for path planning and obstacle avoidance. This pre-map is then refined in the cognition layer to incorporate detailed terrain characteristics and navigability information. In the cognition layer, adaptive behavior is facilitated by integrating a Graph Search Algorithm with a Collision Avoidance System (CAS). This integration optimizes path planning to work in harmony with motion control, ultimately leading to the execution of a seamless trajectory.

\subsection{Phase I  Environment Perception, Self Location, and Data Processing.}
\label{subsec:Phase I}

To initiate autonomous navigation, the robot must recognize its surrounding environment. This involves using different sensors to collect data, which is then processed to create an initial pre-map. Mapping algorithms like occupancy grid mapping \cite{esenkanova2013pre}, SLAM \cite{panigrahi2022localization}, and topological mapping generate this high-level representation of the environment. The pre-map distinguishes navigable and non-navigable areas and, plays a crucial role in subsequent stages, enabling the robot to determine its current location and plan a path to reach its destination.

Since autonomous robots rely on multiple sensors to perceive their environment, it is important to use filtering techniques to merge and refine collected data from these sensors.

Table \ref{Tab:sensors} shows well-established sensors, while table \ref{Tab:filters} shows the most common filters used for data acquisition in environment perception, self-location, data processing, and mapping.

\begin{table}[h]
\caption{Phase I - Well-established sensors.}
    \centering
    \begin{tabular}{p{6.2cm}|p{1.8cm}}
\textbf{Sensors} &  \textbf{Reference}\\ \hline
\multicolumn{2}{p{5cm}}{\textbf{Geo-referencing Systems}}\\\hline
Inertial Navigation System (INS): IMU, Gyroscope, Compass, Altimeter&\cite{he2015polar, panigrahi2022localization}\\\hline 
Attitude and Heading Reference System (AHRS): MEMS Gyroscopes, Accelerometers, Magnetometers.&\cite{urmson2007robust,noh2019decision}\\ \hline
\multicolumn{2}{p{7cm}}{\textbf{Self Location Apparatus (for Dead Reckoning estimation)}}\\\hline
Odometer, Encoder&\cite{li2014multivehicle,iturrate2009noninvasive,snape2011hybrid,li2018algorithm, schultz1998continuous}\\
Optical Encoder&\cite{patle2018path}\\ 
Ultrasonic Sensor & \cite{yang2012efficient} \\ \hline
\multicolumn{2}{c}{\textbf{Eletromagnetic Waves Based Devices}}\\\hline
Radar &\cite{aeberhard2015experience,urmson2007robust,noh2019decision}  \\
Ground Penetrating Radar (GPR)&\cite{he2015polar,urmson2007robust,trautmann2009development,ray2014autonomous}\\
Global Positioning System (GPS) GPS and/or DGPS&\cite{aeberhard2015experience,li2014multivehicle,he2015polar,noh2019decision,patle2018path,yoon2018spline, hajjaj2017bringing}\\\hline
\multicolumn{2}{c}{Cooperative Location Sharing Devices}\\\hline
Automatic dependent surveillance-broadcast (ADS-B), Zigbee, Wireless. &\cite{li2014multivehicle,li2018algorithm}\\
PetriNet Model&\cite{patle2018path}\\ \hline
\multicolumn{2}{c}{Ground beacons based Position Locators Apparatus}\\\hline
Radio Frequency Identification (RFID)&\cite{gueaieb2008intelligent, zhang2018bfvp}\\
Bluetooth wireless&\cite{snape2011hybrid}\\\hline
\multicolumn{2}{c}{Laser Rangefinders (light waves propagation)}\\\hline
 LiDAR  &  \cite{aeberhard2015experience,he2015polar,li2014multivehicle,urmson2007robust,noh2019decision,iturrate2009noninvasive,xiao2015rgb,li2018algorithm,yoon2018spline,patle2018path,macenski2020marathon2, atas2022elevation}\\ 
 Infrared Sensor&\cite{patle2018path, hoy2015algorithms} \\ \hline
 \multicolumn{2}{c}{\textbf{Camera Visual Sensors}}\\\hline
Kinect (Depth), RGB Camera  &\cite{aeberhard2015experience,he2015polar,li2014multivehicle,yan2019real,paul2018application,noh2019decision,snape2011hybrid,yoon2018spline,macenski2020marathon2, atas2022elevation}\cite{xiao2015rgb}
\label{Tab:sensors}
\end{tabular}
\end{table}

\begin{table}[h]
\caption{Phase I - Common filters.}
    \centering
    \begin{tabular}{p{6cm}|p{2cm}}
\textbf{Filters} & \textbf{Reference}\\\hline

\multicolumn{2}{c}{\textbf{Multi Sensor Fusion based Filters}}\\\hline
Kalman Filter&\cite{aeberhard2015experience,li2014multivehicle}\\
Extended Kalman Filter (EKF)&\cite{xiao2015rgb, macenski2020marathon2}\\ \hline

\multicolumn{2}{c}{\textbf{Vision System Filters}}\\\hline 
High Dynamic Range (HDR) Algorithms&\cite{paul2018application} \\ 
Gaussian-based filters&\cite{de2015hybrid, siegwart2011introduction} \\ 
Bayesian-based filters&\cite{zhang2018bfvp, yang2012efficient}
\label{Tab:filters}
    \end{tabular}
\end{table}

This paper does not explore sensor data processing libraries and localization methods. However, it is important to mention that established tools such as YOLO and OpenCV are widely used for detection and localization in the robotics Perception layer \cite{carvalho2022review}. Visual SLAM algorithms \cite{pire2015stereo, ling2017building } and localization methods like Iterative Closest Point (ICP) \cite{atas2021evaluation} and Normal Distributions Transform (NDT) \cite{carvalho2022review} scan matching are also crucial in this layer. Segmentation techniques, utilizing Gaussian-based models \cite{de2015hybrid}, further, enhance localization and visual navigation by accurately classifying navigable paths and reducing spatial requirements for map storage.

\subsection{Phase IIA - Path Planning: Graph Construction}
\label{subsec:Phase II}

Once the perception processing unit in the perception layer extracts meaningful data and creates an initial pre-map using sensor data, the cognition layer refines it. Techniques like surfel-based mapping \cite{atas2022elevation}, Delaunay triangulation, and visibility constraints refine the map into a dense 3D representation \cite{ling2017building}, ensuring accuracy and navigability. This detailed mapping allows the cognition layer to plan precise paths. Table \ref{tab:pp_graph_const} presents usual map-building techniques for autonomous navigation."

\begin{table}[!hbtp]
\centering
\caption{Phase IIA Path Planning: Graph Construction}
\label{tab:pp_graph_const}
\begin{tabular}{p{5.0cm}|p{3.0cm}}
\textbf{Method} & \textbf{Reference}\\\hline

\multicolumn{2}{c}{\textbf{Graph Search Maps}}\\\hline
Voronoi Diagram &\cite{sgorbissa2012planning, gasparetto2015path}\\

Exact Cell Decomposition &\cite{siegwart2011introduction, gasparetto2015path}\\
Height Segmented Map&\cite{karkowski2016real}\\
Surfel-Based Map&\cite{atas2022elevation}\\ \hline 

Approximate Cell Decompositon&\cite{aeberhard2015experience,he2015polar,urmson2007robust,iturrate2009noninvasive, gasparetto2015path}\\
Lattice Graph&\cite{pivtoraiko2009differentially,rufli2009smooth,rufli2009application,ferguson2008motion}\\\hline

\multicolumn{2}{c}{\textbf{Potential Field Maps}}\\\hline

Extended Potential Field Approach&\cite{borenstein1989real,khatib1995extended, gasparetto2015path}\\\hline

\multicolumn{2}{c}{\textbf{Others Methods of Map Building}}\\\hline

Genetic Algorithm (GA)&\cite{li2014multivehicle, gasparetto2015path}\\
CNN Feature Map&\cite{yang2019scene} \\
Spatio-Temporal Voxel Layer (STVL)&\cite{macenski2020marathon2}\\
Dense 3D Mapping & \cite{ling2017building, atas2023navigating, atas2021evaluation}
\end{tabular}
\end{table}

As shown in Table \ref{tab:pp_graph_const}, the Graph Search technique is frequently applied in AMRs, particularly within structured environments like indoor settings. This method effectively utilizes predefined grid or mesh maps for precise navigation.

\subsection{Phase IIB - Path Planning: Graph Search Algorithms}
\label{subsec:PhaseGS}

Once the robot knows its position, environment features, and target, it starts path planning. Roboticists focus on two main parameters in this phase, as noted in \cite{zeng2015survey}:

\begin{itemize}
    \item \textit{completeness}: The ability to find a solution within a finite time.
    \item \textit{optimality}: The ability to compute the most efficient path considering time, energy, or distance. Various strategies for achieving these goals are extensively discussed \cite{gasparetto2015path}.
\end{itemize}

Table \ref{tab:ppGSA} presents well-established path-planning algorithms that have been applied in recent works on autonomous navigation.

\begin{table}[h]
\caption{Phase IIB  Path Planning: Graph Search Algorithms}
    \centering
    \begin{tabular}{p{5.5cm}|p{2.5cm}}
\textbf{Trajectory Generation} &\textbf{Reference}\\ \hline 
\multicolumn{2}{c}{\textbf{Deterministic Graph Search}}\\\hline
Breadth-First Search (BFS)&\cite{siegwart2011introduction, raja2012optimal ,carvalho2022review}\\ 

Depth-First Search (DFS)&\cite{siegwart2011introduction}\\ 

Dijkstra's Algorithm&\cite{casalino2009three,niu2016efficient, carvalho2022review}\\

A* Algorithm&\cite{urmson2007robust,karkowski2016real,sgorbissa2012planning, carvalho2022review}\\

D*Algorithm&\cite{iturrate2009noninvasive, rufli2009application}\\ 

Smac Planner&\cite{macenski2024open}\\ 
\hline

\multicolumn{2}{c}{\textbf{Randomized Graph Search}}\\ \hline

Rapidly Exploring Random Tree (RRT)&\cite{kuwata2009real, atas2021evaluation, carvalho2022review}\\

Spline Sample RRT*&\cite{yoon2018spline}\\

Probabilistic Roadmap&\cite{snape2011hybrid, atas2021evaluation}\\ 
OMPL and SBO Planners &\cite{atas2021evaluation} \\\hline

\multicolumn{2}{c}{\textbf{Derived Algorithms from the previous graph search methods}}\\\hline

Potential Field based Algorithms&\cite{he2015polar,montiel2015path,clemente2018adaptive}\\
Artificial Potential Field (APF)&\cite{li2018algorithm,rostami2018real}\\\hline

Spline Path Planning&\cite{xiao2015rgb}\\
Fuzzy Heuristic Search&\cite{lee2004intelligent, hoy2015algorithms}\\
Firefly Algorithm (FA)&\cite{patle2018path}\\
High Autonomous Driving (HAD) Algorithms&\cite{bahram2014microscopic}\\
Smoothed A* Algorithm&\cite{song2019smoothed}
\label{tab:ppGSA}
    \end{tabular}
\end{table}

For path planning, autonomous vehicles mainly use the A* algorithm and its variants. Traditional methods like BFS and DFS are less preferred due to their inefficiency in optimizing paths. Instead, roboticists develop customized heuristic methods that balance path optimality and computational costs, addressing processing power and decision time constraints\cite{rosa2011fundamentos}.

\subsection{Phase III: Obstacle Avoidance and Trap Landscapes}
\label{subsec:Phase_CASs}

During autonomous operations, mobile robots must navigate around obstacles. If not programmed in the initial phases, collision avoidance systems (CAS) become crucial. These systems are adapted to the robot’s operational environment and kinodynamics, ensuring safe maneuverability. Table \ref{Tab4} details established CAS algorithms, reflecting their varied response times, safety distances, and specific functionalities.

\begin{table}[!hbtp]
\centering
\caption{Phase III:  Obstacle Avoidance and Trap landscape,\;/CASs.}
\label{Tab4}
\begin{tabular}{p{6.3cm}|p{2.0cm}}
\textbf{STRATEGIES  CASs} & \textbf{References}\\\hline

\multicolumn{2}{c}{\textbf{Traditional Algorithms}}\\\hline

Bug Algorithms&\cite{kamon1996new,lumelsky1987path,lumelsky1990incorporating, hoy2015algorithms}\\

Vector Field Histogram (VFH)&\cite{he2015polar,borenstein1989real}\\
VFH+&\cite{ulrich1998vfh+}\\
VFH*&\cite{ulrich2000vfh}\\ 

The Bubble Band Technique&\cite{khatib1997dynamic,urmson2007robust,yoon2018spline}\\
Elastic Band Concept&\cite{quinlan1993elastic}\\

Curvature Velocities Techniques (CVM) &\cite{yang2019scene,simmons1996curvature}\\

Dynamic Windows Approaches&\cite{aeberhard2015experience,minguez2002reactive,fox1997dynamic,brock1999high}\\

The Schlegel Approach &\cite{schlegel1998fast} \\

Nearness Diagram &\cite{minguez2004nearness,minguez2002robot,iturrate2009noninvasive}\\\hline

\multicolumn{2}{c}{\textbf{Virtual Force Field (VFF) Methods}}\\\hline

Gradient Methods&\cite{rostami2018real, konolige2000gradient, hoy2015algorithms}\\
Bacterial Potential Field&\cite{montiel2015path}\\ \hline

\multicolumn{2}{c}{\textbf{Genetic based Algorithms}}\\\hline

Biological Approach&\cite{chen1997crash}\\
Bioinspired Neural Network Algorithm&\cite{cao2016multi, hoy2015algorithms}\\ \hline

\multicolumn{2}{c}{\textbf{Hybrid VFF-Genetic Algorithms}}\\\hline
Evolutionary Behaviour based on Genetic Programming&\cite{clemente2018adaptive}\\\hline

\multicolumn{2}{c}{\textbf{Geometrical Methods}}\\\hline

Boundary Following &\cite{hoy2015algorithms} \\

Collision Cone& \cite{snape2011hybrid} \\

Higher Geometry Maze Routing Algorithm& \cite{raja2012optimal}  \\

Fuzzy / Neurofuzzy Relational Products &\cite{tzafestas1999recent}\\ \hline

\multicolumn{2}{c}{\textbf{Anti-target Approach Laws}}\\\hline

Cone's Geometry-based Calculated Rule&\cite{chakravarthy1998obstacle}
\end{tabular}
\end{table}

Collision Avoidance Systems (CAS) play a vital role in ensuring the safe and effective operation of mobile robots in dynamic environments. By integrating these systems, robots are equipped to dynamically navigate through complex terrains, thereby enhancing their reliability and operational scope for real-world applications.

\subsection{Phase IV: Motion Control And Robot Relocation}
\label{subsec:Phase_control}

To control the movement, speed, position, and orientation of the AMR, various controllers are integrated into the robot, addressing both hardware and software requirements. Vehicles face unique constraints and require specific accuracy and response times due to differing maneuverability capabilities and variable environments. A range of controllers have been developed to meet these needs, as shown in Table \ref{Tab5}.

\begin{table}[h]
\caption{Phase IV Motion Control and Robot Relocation}
    \centering
    \begin{tabular}{p{6cm}|p{2cm}}
\textbf{Controllers}& \textbf{References}\\\hline

\multicolumn{2}{c}{\textbf{Control-Theory Based Controllers}}\\\hline
\multicolumn{2}{c}{Nonlinear Controllers}\\\hline

Time Elastic Band &\cite{macenski2020marathon2}\\ 

Nonlinear Optimal SDRE&\cite{rostami2018real}\\

Pure Pursuit &\cite{kuwata2009real,chen2017pure,macenski2023regulated}\\\hline

\multicolumn{2}{c}{Linear Controllers}\\\hline

Lane Detection and Sliding Mode &\cite{zhang2019autonomous, hoy2015algorithms}\\

PID  (Pose / Velocity)&\cite{he2015polar,urmson2007robust,chen2017pure}\\

Model Predictive Control (MPC) &\cite{hoy2015algorithms, atas2023navigating} \\\hline

\multicolumn{2}{c}{Hybrid Controllers}\\\hline

PID (Pose / Velocity)&\cite{zhang2019autonomous}\\ 

Model Predictive Path Integral Control (MPPI)&\cite{williams2016aggressive}\\\hline

\multicolumn{2}{c}{\textbf{Behaviour Based Controllers}}\\\hline

\multicolumn{2}{c}{Machine Learning}\\\hline

Matlab/hardware Loop &\cite{bahram2014microscopic}\\
Receding Horizon (CNN)&\cite{yang2019scene}\\ \hline

\multicolumn{2}{c}{\textbf{Relocation Techniques And Others Sorts of Controllers}}\\\hline

SLAM &\cite{aeberhard2015experience}\\

Hybrid &\cite{clemente2018adaptive}
\label{Tab5}
\end{tabular}
\end{table}

From Table \ref{Tab5}, it is evident that complex vehicle dynamics and rapidly changing environments necessitate the use of multiple controller types. For example, \cite{kuwata2009real} and \cite{chen2017pure} combined nonlinear controllers with PID controllers to enhance steering and speed regulation. Modern approaches increasingly incorporate nonlinear controllers to address complex differential equations more effectively. Additionally, the use of cooperative predictive controllers has become prevalent, optimizing robot relocation and enhancing motion control efficiency \cite{aeberhard2015experience}.

\subsection{Phase V: Trajectory Execution}
\label{subsec:Phase_Trajectory}

AMRs are equipped with specialized hardware to execute planned routes. Some adhere to predefined trajectories (offline trajectory execution), while others use episodic planning to dynamically integrate planning and execution using sensor data \cite{siegwart2011introduction}. Additionally, AMRs may include real-time replanning modules within the operation layer, eliminating the temporal gap between planning and execution. Behavior trees \cite{macenski2023desks} coordinate diverse planning and execution modules to manage dynamic conditions effectively. This approach increases flexibility, enabling AMRs to adapt to varying environments and operational demands. The distinction between offline and online path planning underscores the necessity for dynamic response capabilities in online scenarios, where unpredictable environments pose significant computational challenges \cite{raja2012optimal}. Table \ref{Tab6} highlights commonly used methods and algorithms for trajectory execution.

\begin{table}[!hbtp]
\centering
\caption{Phase V: Trajectory Execution.}
\label{Tab6}
\begin{tabular}{p{4.5cm}|p{3.5cm}}
\textbf{Method/Algorithm} & \textbf{References}\\\hline

Offline Planning (Dead Reckoning)&\cite{he2015polar,nourbakhsh1999affective,montiel2015path,ray2014autonomous}\\\hline

Episodic Planning (Deferred Planning)&\cite{schultz1998continuous,philippsen2003smooth,arras2002real,urmson2007robust,li2018development,clemente2018adaptive}\\\hline

Integrated Planning and Execution (Real Time Replanning) &\cite{bahram2014microscopic,yang2019scene,rufli2009smooth,rufli2009application,ferguson2008motion,karkowski2016real,noh2019decision,kuwata2009real,hoy2015algorithms,patle2018path,yoon2018spline, macenski2023desks}\\\hline

Hybrid Layers Switching&\cite{antonelli2001real,casalino2009three}
\end{tabular}
\end{table}

\section{Discussion}
\label{Discussion}

The proposed classification of autonomous navigation and the discussion of both well-established and recent techniques and technologies not only contribute to academic understanding but also provide a practical guide for the design and implementation of future mobile robots. This framework also facilitates learning about the various algorithms, components, and sensors that compose autonomous navigation, ranging from perception to operation.

\section{Tendencies}
Recent advancements in Autonomous Mobile Robots (AMRs) underscore trends that enhance their operational capabilities:
\begin{itemize}
    \item \textbf{Enhanced Communication Technologies:} Introduction of 5G technology significantly improves real-time data sharing among AMRs, crucial for industrial automation.
    \item \textbf{Evolution of Navigation Algorithms:} The adoption of evolutionary algorithms and neural networks bolsters real-time decision-making in AMRs.
    \item \textbf{Artificial Intelligence in Cognition:} Integration of deep learning enhances path prediction and decision-making, facilitating advanced cognitive functions.
    \item \textbf{Advanced Sensor Fusion and Data Processing:} Enhanced filtering techniques and new algorithms improve environmental perception and motion planning under challenging conditions.
    \item \textbf{Sensor Diversity and Data Redundancy:} The use of diverse sensors and advanced probabilistic filters increases data reliability, optimizing tasks like SLAM.
    \item \textbf{Nonlinear Filtering Techniques:} Nonlinear filters are critical for accurately processing the complex dynamics of AMRs, enhancing motion control and trajectory prediction.
    \item \textbf{Adaptability Across Environments:} AMRs have advanced in operating across varied environments: terrestrial, aerial, and aquatic, broadening their applications to include disaster response and environmental monitoring.
\end{itemize}
These innovations highlight the dynamic evolution of AMR technology, where flexibility and improved communication are key drivers of enhanced robotic navigation.

\section{Conclusion} 
\label{conclusion}

This paper presented a novel classification system for the domain of autonomous mobile robots (AMRs), aiming to refine the interconnection between diverse technologies for autonomous navigation. By exploring various techniques (such as methodologies, methods, and strategies) and technologies (including sensors, tools and filters) readers can develop deeper understanding of the different phases of autonomous navigation and they can make informed choices of well-established tools for designing mobile robots.


\section*{Acknowledgment}

This research was financed in part by the Coordenação de Aperfeiçoamento de Pessoal de Nível Superior – Brasil (CAPES) – Finance Code 001, and supported by Fundação de Amparo à Pesquisa e Inovação do Estado de Santa Catarina.

\bibliographystyle{IEEEtran} 
\bibliography{refs-cirs} 

\end{document}